\pdfoutput=1
\documentclass[11pt]{article}

\usepackage[]{acl}

\usepackage{times}
\usepackage{latexsym}
\usepackage{url}
\usepackage{multirow}
\usepackage{float}
\usepackage{subfig}
\usepackage[]{xcolor}
\usepackage{diagbox}
\usepackage{makecell}
\usepackage{tikz}
\usepackage{array}
\usepackage{tabularx}
\usepackage{graphicx}
\usepackage{color,array}
\usepackage{comment}
\usepackage{rotating}
\usepackage{lipsum}% http://ctan.org/pkg/lipsum
\usepackage{multicol}
\usepackage{placeins}
\usepackage{adjustbox} %smn
\usepackage{makecell} %smn
\usepackage{arydshln} %smn
\PassOptionsToPackage{bookmarks=false}{hyperref}
\usepackage{latexsym}
\usepackage{nicematrix}
\restylefloat{table}
\usepackage{latexsym}

\usepackage[T1]{fontenc}
\usepackage[utf8]{inputenc}

\usepackage{microtype}

\definecolor{backgG}{RGB}{255, 255, 153}
\definecolor{tagtxtG}{RGB}{102, 102, 0}
\definecolor{backgPc}{RGB}{179, 255, 179}
\definecolor{tagtxtPc}{RGB}{0, 102, 0}
\definecolor{backgPw}{RGB}{255, 179, 179}
\definecolor{backgPw}{rgb}{0.0, 1.0, 1.0}
\definecolor{tagtxtPw}{RGB}{0.0, 1.0, 1.0}
\definecolor{backgPo}{rgb}{0.0, 1.0, 1.0}
\definecolor{tagtxtPo}{RGB}{102, 0, 0}
\definecolor{backgPm}{rgb}{0.98, 0.81, 0.69}
\definecolor{tagtxtPm}{RGB}{0,1,1}

\title{Named Entity Recognition Based Automatic Generation of Research Highlights}
\author{Tohida Rehman \\
  Jadavpur University, India \\
  \texttt{tohida.rehman@gmail.com} \\ \And
  Debarshi Kumar Sanyal \\
  Indian Association for the 
  Cultivation of Science\\% India \\ 
  \texttt {debarshisanyal@gmail.com} \\ \AND
  Prasenjit Majumder \\
  TCG CREST, India \\
  \texttt{prasenjit.majumder@gmail.com} \\
   \And
  Samiran Chattopadhyay \\
  TCG CREST; Jadavpur University, India \\ 
  \texttt{samirancju@gmail.com} \\
}

\begin{document}
\maketitle
\begin{abstract}
A scientific paper is traditionally prefaced by an abstract that summarizes the paper. Recently, research highlights that focus on the main findings of the paper have emerged as a complementary summary in addition to an abstract. However, highlights are not yet as common as abstracts, and are  absent in many papers.
In this paper, we aim to automatically generate research highlights using different sections of a research paper as  input. We investigate whether the use of named entity recognition on the input improves the quality of the generated highlights. In particular, we have used two deep learning-based models: the first is a pointer-generator network, and the second augments the first model with coverage mechanism. We then augment each of the above models with named entity recognition features. 
The proposed method can be used to produce highlights for papers with missing highlights. Our experiments show that adding named entity information improves the performance of the deep learning-based summarizers in terms of ROUGE, METEOR and BERTScore measures.
\end{abstract}

\section{Introduction}
\label{intro}
Every research domain has an overabundance of textual information, with new research articles published on a daily basis. The number of scientific papers is increasing at an exponential rate \citep{bornmann2021growth}. According to reports, the number of scientific articles roughly doubles every nine years \citep{van2014global}. For a researcher, keeping track of any research field is extremely difficult even in a narrow sub-field. Nowadays, many publishers request authors to provide a bulleted list of research highlights along with the abstract and the full text. It can help the reader to quickly grasp the main contributions of the paper.\\
Automatic text summarization is a process of shortening a document by creating a gist of it. It encapsulates the most important or relevant information from the original text. 
%When it comes to information extraction, scientific documents have different requirements due to their compact, imprecise discourse style \citep{biber2010challenging} than news articles. 
Scientific papers are generally longer documents than news stories and have a different discourse structure. 
%including more number of OOV terms. 
Additionally, there are less resources available on scholarly documents to train text summarization systems. There are two broad approaches used in automatic text summarization \citep{luhn1958automatic,radev2002introduction}: Extractive summarization and abstractive summarization. Extractive summarization generally copies whole sentences from the input source text and combines them into a summary, discarding irrelevant sentences from the input \citep{jing2000cut,knight2002summarization}. But recent trends use abstractive summarization which involves natural language generation to produce novel words and capture the salient information from the input text \citep{rush2015neural,nallapati2016abstractive}. 
%In general, abstractive summarization attempts to cover all of the important salient points of a document in a concise and coherent manner. 
Our aim is to generate research highlights from a research paper using an abstractive approach.
But an abstractive summarizer using a generative model like a pointer-generator network \cite{See2017GetTT} sometimes generates meaningless words in the output. In particular, for named entities which are multi-word strings, incorrect generation of a single word within the string (e.g., suppose  instead of generating `artificial neural network', it generates `artificial network') may corrupt the meaning of the whole entity and its parent sentence. So we propose to perform named entity recognition (NER) on the input and treat a named entity as a single token before the input passes through the summarizer. This will avoid their fragmentation in the output.

The main contributions of this paper are:
\begin{enumerate}
    \item We propose a mechanism to combine named entity recognition with pointer-generator networks having  coverage mechanism to automatically generate research highlights, given the abstract of a research paper. 
    To the best of our knowledge, this work is the first attempt to use NER in pointer-generators with coverage mechanism \citep{See2017GetTT} to generate research highlights. 
    \item We analyze the performance of generating research highlights for the following different input types: (a) the input is the abstract only, (b) the input comprises the abstract and the conclusion, (c) the input comprises the introduction and the conclusion.
    \item We evaluate the performance of the models using ROUGE \citep{lin2004rouge}, METEOR \citep{banerjee2005meteor}, and BERTScore \citep{zhang2019bertscore} metrics.
\end{enumerate}

\section{Literature survey}
Early works on summarization of scientific articles include  \citep{kupiec1995trainable} where an extractive summarization technique is proposed and evaluated on a small dataset of 188 scientific documents, and \citep{teufel-moens-2002-articles} which exploits the rhetorical status of assertions to summarize scientific articles. 
More recently, Lloret et al. \citep{lloret2013compendium} have developed a new corpus of computer science papers from \texttt{arXiv.org} that contains pairs of (Introduction, Abstract). An approach proposed to generate abstracts from a research paper using Multiple Timescale model of the Gated Recurrent Unit (MTGRU) may be found in \citep{kim-etal-2016-towards}.
Surveys on summarization of scholarly documents appears in \citep{altmami2020automatic}, \citep{el2021automatic}.
Generating research highlights from scientific articles is not the same as document summarization. A supervised machine learning approach is proposed in \citep{collins2017supervised} to identify relevant highlights from the full-text of a paper using a binary classifier. They have also contributed a new benchmark dataset containing author written research highlights for more than 10,000 documents. All documents adhere to a consistent discourse structure. Instead of a simple binary classification of sentences as highlights or not, \citep{cagliero2020extracting} used multivariate regression methods to select the top-$K$ most relevant sentences as research highlights.
\citep{hassel2003exploitation} proposed a method to use appropriate weight for the named entity tagger into the SweSum summarizer for Swedish newspaper texts. \citep{marek2021text} proposed an extractive summarization technique that determines a sentence's significance based on the density of named entities.
\citep{rehman2021automatic} used a pointer-generator model with coverage  \citep{See2017GetTT} to generate research highlights from abstracts. The present work, unlike the existing ones, uses NER to avoid incorrect phrases from being generated by the decoder. Note that pretrained summarization models like PEGASUS \citep{zhang2020pegasus}, T5 \citep{raffel2019exploring}, and BART \citep{devlin2018bert}  are trained on generic texts. Fine-tuning them to a specific (e.g., scientific)  domain requires large memory and computational resources; in this context, this paper provides a simpler alternative.  

\section{Methodology}
We use a pointer-generator network \citep{See2017GetTT} as our baseline model. While the pointer-generator model \citep{See2017GetTT} first tokenizes a document using Stanford CoreNLP %\footnote{\url{https://stanfordnlp.github.io/CoreNLP/}}
tokenizer 
and converts the tokens to word embeddings (trained with the model), the method we propose here performs NER on the input document and considers a named entity as a single token when training the model. 
We perform experiments with 4 variants: (1) the original pointer-generator model proposed in \citep{See2017GetTT} (PGM), (2) pointer-generator model integrating coverage mechanism (proposed in \citep{tu2016modeling}) (PGM + Cov), described in the same work \citep{See2017GetTT}, (3) NER-based pointer-generator model (NER + PGM), and (4) NER-based pointer-generator model with coverage mechanism (NER + PGM + Cov).

\subsection{NER-based pointer-generator network}
This model consists of an NER-based tokenizer layer and a pointer-generator network. 
The NER-based tokenizer layer converts the words in the input document to a sequence of tokens, thus preserving an entity name as a single token. In particular, it uses the named entity recognizer in  spaCy\footnote{\url{https://spacy.io/usage/linguistic-features.}} However, we do not use entity types. 
We do not use pretrained word embeddings as \citep{nallapati2016abstractive} do; in our case token embeddings are learned from scratch during training. 
Here, the main role of NER is that instead of directly feeding the normal tokens of the input document into the encoder, we are passing the NER-based tokens.

\begin{comment}
\begin{figure*}[!htb] 
\centering 
\includegraphics[width=16cm,height=8cm]{Model_diagram_NER}
\caption{Proposed model: NER-based pointer-generator network with coverage mechanism.}
\centering
\label{fig:Model_diagram_NER}
\end{figure*}
\end{comment}

\begin{table*}[!htbp]
    \centering
    \begin{adjustbox}{width=.99\linewidth}
    {\begin{tabular}{llccccc}  \hline
%& &{ROUGE-1} &{ROUGE-2} &{ROUGE-L} &METEOR & \\ 
 Input &Model Name &ROUGE-1 &ROUGE-2 &ROUGE-L &METEOR &BERTScore \\ \hline
\multirow{4}{4em}{abstract only} &PGM &35.44 &11.57 &29.88 &25.4 &83.80 \\ 
&PGM + Cov &36.57 &12.3 &30.69 &25.4 &84.05 \\ 
&NER + PGM &35.88 &12.78 &33.21 &29.14 &86.02 \\ 
&NER + PGM + Cov &\bf{38.13} &\bf{13.68} &\bf{35.11} &\bf{31.03} &\bf{86.3} \\ \hline
\multirow{4}{4em}{abstract + conclusion} &PGM &29.85 &8.16 &25.80 &19.38 &83.19\\ 
&PGM + Cov &31.70 &8.31 &26.72 &20.92 &83.49 \\ 
&NER + PGM &35.12 &12.37 &32.37 &28.34 &86.08 \\ 
&NER + PGM + Cov &\bf{37.48} &\bf{13.26} &\bf{34.95} &\bf{28.97} &\bf{86.64}\\ \hline
\multirow{4}{4em}{introduction + conclusion} &PGM &29.78 &7.47 &25.15 &19.25 &83.05 \\ 
&PGM + Cov &31.63 &7.65 &26.25 &20.24 &83.32 \\ 
&NER + PGM &31.74 &9.18 &29.44 &23.82 &85.78 \\ 
&NER + PGM + Cov &\bf{34.24} &\bf{9.82} &\bf{31.92} &\bf{25.36} &\bf{86.1} \\ \hline
\end{tabular} 
}
\end{adjustbox}
\vspace{4mm}
\caption{Evaluation of pointer-generator type models: F1-scores for ROUGE, METEOR and BERTScore on various inputs from CSPubSumm dataset. All our ROUGE scores have a 95\% confidence interval of at most $\pm$ 0.25 as reported by the official ROUGE script.} \label{Table:par_all_types_rouge_meteor_bert}
\vspace{-6mm}
\end{table*}
\section{Experimental setup}
\subsection{Data sets}
We use the data sets published by Collins et al.  \citep{collins2017supervised}, which contains URLs of 10147 computer science publications from ScienceDirect (\texttt{https://www.sciencedirect.com/}).
Title, abstract, author-written research highlights, a list of keywords referenced by the authors, introduction, related work, experiment, conclusion, and other important subsections as found in typical research papers are all included for each document. In our setup, every example in this data set is organised as follows: \textit{(abstract, author-written research highlights, introduction, and conclusion)}. We use 8116 examples for training, 1017 examples for validation, and 1014 examples for testing.

\subsection{Data processing}
We have removed digits, punctuation, and special characters from the documents and lowercased the entire corpus. The \texttt{retokenizer.merge} method of spaCy is used to tokenize and merge several tokens into one single token based on the named entities in the document. Instead of individual tokens of ``artificial", ``neural", and ``network", we pass all the three  tokens together as a single token ``artificial neural network"  (referenced as vocab index 17). 
The data set is then reorganized in several ways to conduct various experiments. We organize the data set as \textit{(abstract, author-written research highlights)},  \textit{(abstract + conclusion, author-written research highlights)}, and  \textit{(introduction + conclusion, author-written research highlights)}  where `+' denotes text concatenation. Since abstract and introduction usually emphasize the same aspects of the paper, we have not included them together.
%Normally the main findings of a research paper are highlighted in the abstract and introduction. Therefore, we have considered these sections for our experiments. 
%Main contribution of a paper are covered in the abstract as well as introduction section, since we do not use the (abstract + introduction)  as an input rather we use (abstract + conclusion), and (introduction + conclusion).
%%\textcolor {red}{
In this data set, the average abstract length is 186 tokens, while the average author-written-highlight length is 52 tokens. When we considered abstract and conclusion, the average length was 643. When we considered introduction and conclusion, the average length was 1234. 
Therefore in our model, we have set the maximum number of input tokens to 400 when the abstract is taken as the input. For all other cases, the maximum count of input tokens is set to 1500. In all cases, the generated research highlights have a maximum token count of 100. 
We trained all models on Tesla V100-SXM2-16GB \texttt{Colab Pro+} that supports GPU-based training. For all models, we used two bidirectional LSTMs with cell size of 256, word embeddings of dimension 128, and maximum vocabulary size of 50K tokens. We considered gradient clipping with a maximum gradient norm of 1.2. We use other hyperparameters as suggested by \citep{See2017GetTT}.

\subsection{Comparison with previous works}

Table \ref{Table:per_comparison_others} compares the performance of our proposed approach (NER + PGM + Cov) with other competitive baselines in the literature, on the CSPubSumm data set in terms of the ROUGE-L F1-score.
\begin{table}[!htb]
\centering
\begin{adjustbox}{width=1.0\linewidth}
\begin{tabular}{p{6cm}p{1cm}}  \hline
{Model Name} &ROUGE-L (F1) \\ \hline
LSTM \citep{collins2017supervised} &29.50\\ 
GBR \citep{cagliero2020extracting} &31.60 \\ 
PGM + Cov + GloVe \citep{rehman2021automatic} &29.14 \\
NER + PGM + Cov (ours) &\bf{35.11} \\ \hline
\end{tabular} 
\end{adjustbox}
\caption{Comparison of ROUGE-L (F1-score) of the proposed model with that of the other approaches. \label{Table:per_comparison_others}}
\end{table}

\begin{figure*}[!htb] 
\centering 
\small
\begin{tabular}{ |p{15.5cm}|} \hline
{\bf Author-written research highlights:} This paper studies an image collection planning problem for a korean satellite kompsat 2( korea multi purpose satellite 2). 
We present a binary integer programming model for this problem in a multi orbit long term planning environment. 
A heuristic solution approach based on the lagrangian relaxation and subgradient methods is provided. \\\hline
{\bf PGM:} Image collection with \colorbox{backgG}{2 korea 2 image acquisition [UNK]} of \colorbox{backgG}{segment square lengths} as well as well as \colorbox{backgG}{collection relaxation}. Integer programming model exploiting lagrangian relaxation. Results of our computational experiment based on the lagrangian relaxation and subgradient methods. \\\hline 	    
{\bf PGM + Cov:} Image collection planning problem with a multi orbit multi purpose satellite 2 korea multi purpose. A binary integer programming model for a multi orbit long term planning. A heuristic solution approach based on the lagrangian relaxation and subgradient methods. \colorbox{backgPm}{A heuristic solution approach based on the lagrangian relaxation and subgradient methods}.\\ \hline
{\bf NER + PGM:} We present a binary integer programming model for this image collection planning problem for a \colorbox{backgPc}{korean satellite} kompsat.  
We present a binary integer programming model for this problem in a multi orbit long term planning environment.
Combining the lagrangian relaxation and subgradient methods using the \colorbox{backgPm}{lagrangian relaxation and subgradient methods} to solve the top problem.\\\hline
{\bf NER + PGM + Cov:} We present an image collection planning problem for a \colorbox{backgPc}{korean satellite} kompsat.
We present a binary integer programming model for image collection planning. We show the heuristic approach based on the lagrangian relaxation. We present the results on a multi orbit long term planning environment.\\ \hline
\end{tabular} 	
\caption{Input is only an abstract from CSPubSumm data set. Highlights produced by each of the four models are shown.  Input and author-written research highlights taken  \texttt{https://www.sciencedirect.com/science/article/pii/S037722171300307X}}	
\label{fig:sample_Abs_RHS} 
\end{figure*}
\section{Results}
%\vspace{-2.5mm}
\subsection{Comparison of pointer-generator type models}
In this sub-section, we report the results of experiments on the CSPubSumm data set for various input types. 
Table \ref{Table:par_all_types_rouge_meteor_bert} shows the F1-scores for ROUGE-1, ROUGE-2, ROUGE-L, METEOR and BERTScore metrics for various inputs from the test dataset. Among the four models, the NER-based pointer-generator network with coverage mechanism achieves the highest ROUGE, METEOR and BERTScore values.  It appears that treating an entity as a single token in the input helps to learn better embeddings and results in more controlled generation of the output, thereby reducing semantically invalid words and phrases. We aim to investigate this aspect in future. The (NER + PGM + Cov) model achieves the highest scores  when the input is the abstract, indicating that most of the findings reported by the research highlights are already in the abstract, and  adding additional sections to the input contributes to noise for the model.
\subsection{Case study}
Figure \ref{fig:sample_Abs_RHS}, \ref{fig:sample_Abs_Con_RHS} and \ref {fig:sample_int_Con_RHS} shows sample outputs generated by 
our models for various input types. 
In all the case studies reported below, we highlight examples of  \colorbox{backgG}{factual errors},  \colorbox{backgPm}{repeating words} and \colorbox{backgPc}{correct named entities}.
Note that the NER-based models correctly generate the named entities in the output.

\subsection{Manual evaluation}
We selected a set of 25 papers, their author-written highlights (\textbf{A}) and their highlights from only the (NER + PGM + Cov) model (\textbf{M}). We recruited 16 human annotators (possessing or pursuing advanced degrees in software engineering at premier universities in India) to independently rate a given summary on a scale of 1(low) to 5(high) for adequacy and fluency (separately). Each rater was given the full text of a paper and \textit{either} the author-written \textit{or} the machine-generated highlights of the paper, but not told which one. Each summary was rated independently by two raters. On fluency, the average score for \textbf{A} was 4.02 and that for \textbf{M} was 3.3, while on adequacy, the average score for \textbf{A} was 3.82 and that for \textbf{M} was 3.12. This shows the machine-generated highlights are only slightly worse than the human-written ones.

\section{Conclusion}
We applied four different deep neural models to generate research highlights from a research paper. The NER-based pointer-generator model with coverage mechanism achieved the best performance in all cases. In future, we intend to investigate in detail why NER improves the generated highlights. We also aim to ensure that the generated highlights are correct in syntax and semantics.

\begin{figure*}[!tbp] 
\centering 
\begin{tabular}{ |p{15.5cm}|} \hline
{\bf Author-written research highlights:} We present a lightweight non parametric approach to generate wrinkles for 3d facial modeling and animation. Our method represents a convenient approach for generating plausible facial wrinkles with low cost. Our method enables the reconstruction of captured expressions with wrinkles in real time. \\\hline
{\bf PGM:} We propose a non parametric facial modeling 3d face models from the \colorbox{backgPm}{3d facial modeling}. Synthesize the 3d face expression model with 3d depth camera is considered. Propose a non parametric \colorbox{backgG}{face method to evaluate the performance of the 3d facial modeling}. Method is provided to evaluate the performance of the proposed approach with respect to the existing method. \\\hline 	    
{\bf PGM + Cov:} We propose non parametric face acquisition 3d facial modeling models. Face expression model is based on the texture synthesis of multiple subjects. Synthesis method guarantees to 3d face \colorbox{backgPm}{face} acquisition. One high quality 3d face model is studied.\\ \hline
{\bf NER + PGM:} A novel synthesis method is proposed to enhance the wrinkles using a single low cost kinect camera. The lightweight feature of the method is that it can generate plausible wrinkles using a single low cost kinect camera and one high quality 3d face model with details as the example. User specific expressions are used as blendshapes to capture facial animations in real time kinect camera and one high quality 3d face model with details.\\\hline
{\bf NER + PGM + Cov:} We present a lightweight non parametric method to generate wrinkles for 3d facial modeling and animation. The lightweight feature of the method is that it can generate plausible wrinkles. Our method is low cost and convenient for common users. \\ \hline
\end{tabular} 	
\caption{Input is (abstract + conclusion) from CSPubSumm data set. Highlights produced by each of the four models are shown.  Input and author-written research highlights taken  \texttt{https://www.sciencedirect.com/science/article/pii/S0010448514001857}}	
\label{fig:sample_Abs_Con_RHS} 
\end{figure*}

\begin{figure*}[!tbp] 
\centering 
\begin{tabular}{ |p{15.5cm}|} \hline
{\bf Author-written research highlights:} We propose a novel parallel 3d delaunay meshing algorithm for large scale simulations. The model information is kept during parallel triangulation process. A 3d local non delaunay mesh repair algorithm is proposed. The meshing results can be very approaching to the model boundary. The method can achieve high parallel performance and perfect scalability.  \\\hline
{\bf PGM:} We propose a solid model boundary preserving method for large scale parallel 3d delaunay meshing. Parallel 3d local mesh \colorbox{backgPm}{3d} delaunay meshing algorithm is proposed. Mesh reconstruction is iteratively performed to meet \colorbox{backgG}{both the mesh and the shared interfaces}. Propose a parallel 3d local mesh reconstruction algorithm to \colorbox{backgG}{construct delaunay triangulation}. Results show high performance and perfect scalability. \\\hline 	    
{\bf PGM + Cov:} A new semantic parallel algorithm is proposed for large scale parallel 3d delaunay meshing. \colorbox{backgG}{Numerical local mesh is the sampling vertices} for the problem 3d delaunay meshing. Propose a parallel su based partitioning algorithm by \colorbox{backgG}{solving the algorithm}. Proposed algorithm is highly parallelized to large scale sets and high quality partition walls.\\ \hline
{\bf NER + PGM:} We propose a solid model boundary preserving method for large scale 3d delaunay meshing. The 3d boundary representation model information is kept during the entire parallel 3d delaunay triangulation process. \colorbox{backgPm}{The 3d boundary representation model information is kept during the entire parallel 3d delaunay} \colorbox{backgPm}{triangulation process}. A parallel 3d local mesh optimization algorithm is presented. Experimental results demonstrate high performance and perfect scalability.\\\hline
{\bf NER + PGM + Cov:} We propose a solid model boundary preserving method for large scale parallel delaunay meshing. The 3d boundary representation model information is during the entire parallel 3d delaunay triangulation process. A parallel local mesh refinement algorithm to repair the non delaunay mesh is proposed. A parallel 3d delaunay mesh refinement is presented. Experimental results demonstrate scalability performance. \\ \hline
\end{tabular} 	
\caption{Input is (introduction + conclusion) from CSPubSumm data set. Highlights produced by each of the four models are shown.  Input and author-written research highlights taken  \texttt{https://www.sciencedirect.com/science/article/pii/S0010448514001821}}	
\label{fig:sample_int_Con_RHS} 
\end{figure*}

\FloatBarrier
\bibliography{anthology,custom}
\bibliographystyle{acl_natbib}
\end{document}